\documentclass{article}
\usepackage{spconf,amsmath,graphicx,amssymb}
\usepackage{hyphenat}
\usepackage{verbatim}
\usepackage{subfigure,float}
\usepackage{multirow}
\usepackage{xr-hyper}
\usepackage{hyperref}
\usepackage[]{algorithm2e}

\title{LEARN-BY-CALIBRATING: USING CALIBRATION AS A TRAINING OBJECTIVE}

\name{Jayaraman J. Thiagarajan$^{\dagger}$\thanks{This work was performed under the auspices of the U.S. Department of Energy by Lawrence Livermore National Laboratory under Contract  DE-AC52-07NA27344.} , Bindya Venkatesh$^{\ddagger}$ and Deepta Rajan$^*$}
\address{$^{\ddagger}$ Lawrence Livermore National Labs, $^{\dagger}$Arizona State University, $^*$IBM Research AI}
\begin{document}
%
\maketitle
\begin{abstract}
Calibration error is commonly adopted for evaluating the quality of uncertainty estimators in deep neural networks. In this paper, we argue that such a metric is highly beneficial for training predictive models, even when we do not explicitly measure the uncertainties. This is conceptually similar to heteroscedastic neural networks that produce variance estimates for each prediction, with the key difference that we do not place a Gaussian prior on the predictions. We propose a novel algorithm that performs simultaneous interval estimation for different calibration levels and effectively leverages the intervals to refine the mean estimates. Our results show that, our approach is consistently superior to existing regularization strategies in deep regression models. Finally, we propose to augment partial dependence plots, a model-agnostic interpretability tool, with expected prediction intervals to reveal interesting dependencies between data and the target.
\end{abstract}
\begin{keywords}
deep regression, calibration, prediction intervals, partial dependence plot, uncertainty quantification
\end{keywords}
\section{Introduction}
\label{sec:intro}
Machine learning (ML) techniques, such as deep neural networks, have led to incredible advances in a wide variety of applications. As the complexity of ML techniques increases, the black-box nature of these data-centric approaches has become a critical bottleneck. The inability to obtain a holistic understanding of strengths and weaknesses of models makes it challenging to deploy such models in the real-world. For example, a classifier model can often produce highly concentrated softmax probabilities -- suggesting a reliable class assignment -- even for out-of-distribution test samples. The intricate interactions between data sampling, model selection and the inherent randomness in complex systems strongly emphasize the need for a rigorous characterization of ML algorithms~\cite{gal2016uncertainty}. In conventional statistics, uncertainty quantification (UQ) provides this characterization by measuring how accurately a model reflects the physical reality and by studying the impact of different error sources on the prediction~\cite{smith2013uncertainty}. Consequently, several recent efforts have proposed to utilize prediction uncertainties in deep models to shed light onto when and how much to trust the predictions~\cite{healthuqleveraging,uqbioopportunities,cvuncertainties,levasseur2017uncertaintiesphysics}. These uncertainty estimates can also be used for enabling safe ML practice, e.g., identifying out-of-distribution samples, detecting anomalies/outliers, delegating high-risk predictions to experts, defending against adversarial attacks etc.

Existing deep uncertainty estimation methods~\cite{gal2016dropout,gal2017concrete,lakshminarayanan2017ensemble,ghahramani2015probabilistic} attempt to accumulate uncertainties that might arise from modeling and data statistics~\cite{heskes1997practical}. By incorporating uncertainties into training, we expect the predictive models to generalize better. It is typical to evaluate these uncertainties by constructing prediction intervals and evaluating their calibration: An interval is well calibrated if the likelihood of the true target falling in the interval is consistent with the confidence level of the interval. Many approaches place specific statistical priors on predictions or on hidden units in a network, and perform parameter inferencing. While this assumption has enabled tractable inference, it is often found that the resulting intervals are not inherently well calibrated~\cite{kuleshov2018accurate}.

\paragraph*{Proposed Work:}In this paper, we explore the use of calibration as a learning objective while building deep models for regression tasks. More specifically, we adopt a black-box approach, wherein we do not explicitly measure the uncertainties, but directly construct prediction intervals with the objective of minimizing calibration error. Our approach performs alternating optimization between a mean estimator network, and an interval estimator that simultaneously estimates intervals for multiple calibration levels. Surprisingly, we find that calibration is an effective objective for producing mean estimators that are highly accurate, which we demonstrate using empirical studies. More importantly, though our approach is black-box in nature and does not utilize tractable priors~\cite{gal2016uncertainty}, it still produces prediction intervals that achieve significantly lower calibration error, when compared to existing methods. Finally, we propose to augment partial dependence plots~\cite{friedman2001greedy}, which are routinely used to study marginal dependence in regression models, with prediction intervals.

\section{Learn-By-Calibrating}
\label{sec:motivation}
We begin by motivating the use of calibration as a training objective in predictive models and subsequently describe an algorithm that relies on simultaneous prediction interval estimation for different calibration levels. The notion of calibration comes from the uncertainty quantification literature~\cite{smith2013uncertainty}. Since the focus of this paper is on regression tasks, we formally define calibration in that context. Let us assume that a model $\mathcal{F}$ that takes in $\mathrm{x} \in \mathbb{R}^d$ as input, produces the prediction for a target variable $\mathrm{y} \in \mathbb{R}$, along with an interval, i.e., $[\hat{\mathrm{y}} - \mathrm{\delta}, \hat{\mathrm{y}} + \mathrm{\delta}]$, where $\hat{\mathrm{y}}$ is the mean estimate and $2\times\mathrm{\delta}$ is the predicted interval width. Note, while the mean estimate is a random variable, an interval estimate is a random interval. While an interval does or does not contain a certain value, a random interval has a certain probability of containing a value. Suppose that $p[\hat{\mathrm{y}} - \mathrm{\delta} \leq \mathrm{y} \leq \hat{\mathrm{y}} + \mathrm{\delta}] = \alpha$, where $\alpha \in [0,1]$, then the random interval is referred to as a $100 \times \alpha \%$ confidence interval. The intervals produced by $\mathcal{F}$ are considered to be well calibrated if the probability of the true target falling in the interval matches the true empirical probability.

Though the idea of calibration has been widely adopted for evaluating the quality of uncertainty estimators in deep learning~\cite{gal2016uncertainty}, we argue that it is an effective choice for designing loss functions in deep regression. In general, the optimization objective for predictive modeling can be written as
\begin{equation}
    \displaystyle \sum_i \rho\bigg(\mathrm{y}_i - \mathcal{F}(\mathrm{x}_i; \Theta)\bigg) + \lambda \mathcal{R}(\Theta),
\end{equation}where $\rho$ denotes a loss function that measures the discrepancy between the true targets $\mathrm{y}$ and the predictions $\mathcal{F}(\mathrm{x})$, $\Theta$ are the model parameters and $\mathcal{R}$ is a suitable regularization. The loss function $\rho$ is chosen based on assumptions on the structure of the residual $\mathrm{y} - \mathcal{F}(\mathrm{x})$. For example, when the elements of the residual are assumed to follow a Gaussian distribution, it is useful to impose the $\ell_2$ penalty. However, restricting $\mathcal{F}$ to produce only point estimates limits our ability in characterizing the confidence of $\mathcal{F}$ on its predictions. Consequently, in practice, it is often beneficial to consider prediction distributions or intervals in lieu of point estimates. For example, heteroscedastic neural networks (HNN) place a Gaussian prior on the prediction at each sample and optimize for $\Theta$ such that the (Gaussian) likelihood of the prediction distribution containing the true target is maximized. While identifying the uncertainty sources that the prediction intervals from a HNN actually capture is known to be challenging, the heteroscedastic regression objective is nevertheless a flexible loss function for learning. In this spirit, we propose to utilize calibration as a training objective in deep regression models, wherein no specific prior assumptions (e.g. Gaussianity) are required on the predictions or the residuals.

\section{Algorithm}
\label{sec:algorithm}
\RestyleAlgo{boxruled}
\begin{algorithm}[t]
	
	\KwIn{Labeled data $\{(\mathrm{x}_i, \mathrm{y}_i)\}_{i=1}^N$, iterations $T$, set of calibration levels $A$.}
	\KwOut{Trained models $\mathcal{F}$ and $\mathcal{G}$}
	\textbf{Initialization}:Randomly initialize the model parameters \;
	
	\For{$T$ iterations}{
	    Randomly select a level $\alpha$ from $A$ \;
	    \vspace{0.1in}
	    /*\textit{For fixed $\mathcal{F}$:}*/ \\
	    Compute widths $\mathrm{\delta_i}^{\alpha} = \mathcal{G}(\mathrm{x}_i) \quad \forall i = 1 \cdots N$ \;
	    Estimate the loss function $\mathrm{L}_{\mathcal{G}}$ using Eq. (\ref{eqn:ece}) \;
	    Update parameters $\Phi^* = \arg \min_{\Phi} \mathrm{L}_{\mathcal{G}}$ \;
	    
	    \vspace{0.1in}
	    /*\textit{For fixed $\mathcal{G}$:}*/ \\
	    Compute mean $\hat{\mathrm{y}}_i = \mathcal{F}(\mathrm{x}_i) \quad \forall i = 1 \cdots N$ \;
	    Estimate the loss function $\mathrm{L}_{\mathcal{F}}$ using Eq. (\ref{eqn:hinge}) \;
	    Update parameters $\Theta^* = \arg \min_{\Theta} \mathrm{L}_{\mathcal{F}}$ \;
		}
	\textbf{return} $ \Theta^*$, $\Phi^*$
	\caption{Learn-by-Calibrating}\label{algo-d}
\end{algorithm}

We now outline the algorithm for calibration-driven learning in regression. In our formulation, we consider a model $\mathcal{F}: \mathrm{x} \mapsto \mathrm{y}$ with parameters $\Theta$ to produce a mean estimate $\hat{\mathrm{y}}$, and a width estimator $\mathcal{G}: \mathrm{x} \mapsto \mathbb{R}^+$ with parameters $\Phi$ to produce $\mathrm{\delta}$, jointly returning the interval $[\hat{\mathrm{y}}-\mathrm{\delta}, \hat{\mathrm{y}}+\mathrm{\delta}]$. Since we do not assume the predictions to follow a Gaussian distribution, the widths corresponding to each empirical calibration level $\alpha$ cannot be implicitly evaluated. Hence, we allow the width estimator $\mathcal{G}$ to simultaneously produce widths, $\mathrm{\delta}^{\alpha}$, for a pre-defined range of $\alpha$'s (example, $[0.1, 0.3, 0.5, 0.7, 0.9]$) in an attempt to obtain intervals for increasing levels of calibration. 

\begin{table*}[]
\centering
\renewcommand{\arraystretch}{1.3}
\caption{Performance evaluation of predictive models inferred with the calibration objective. For comparison, we report results from popular baselines that are equipped with uncertainty estimators.}
\vspace{0.05in}
\begin{tabular}{|c|c|c|c|c|c|c|c|c|c|c|}
\hline
\multirow{2}{*}{\textbf{Dataset}} & \multicolumn{2}{c|}{\textbf{MC Dropout}~\cite{gal2016dropout}} & \multicolumn{2}{c|}{\textbf{Concrete Dropout}~\cite{gal2017concrete}} & \multicolumn{2}{c|}{\textbf{BNN}~\cite{ghahramani2015probabilistic}}                                      & \multicolumn{2}{c|}{\textbf{HNN}~\cite{gal2016uncertainty}} & \multicolumn{2}{c|}{\textbf{Proposed}} \\ \cline{2-11} 
                                  & \textbf{RMSE}       & \textbf{ECE}       & \textbf{RMSE}          & \textbf{ECE}          & \multicolumn{1}{l|}{\textbf{RMSE}} & \multicolumn{1}{l|}{\textbf{ECE}} & \textbf{RMSE}    & \textbf{ECE}   & \textbf{RMSE}      & \textbf{ECE}      \\ \hline
Crime                             & 0.15                & 0.81               & 0.14                   & 0.91                  & 0.16                               & 0.72                              & 0.14             & 0.58           & \textbf{0.13}      & \textbf{0.11}     \\ \hline
Red Wine                          & 0.64                & 0.39               & 0.68                   & 0.86                  & 0.79                               & 0.46                              & 0.65             & 0.19           & \textbf{0.6}      & \textbf{0.06}     \\ \hline
White Wine                        & 0.75                & 1.07               & 0.79                   & 0.94                  & 0.83                               & 1.14                              & 0.77             & 1.09           & \textbf{0.72}      & \textbf{0.06}     \\ \hline
Parkinsons                        & 4.33                & 1.09               & 4.88                   & 1.22                  & 5.49                               & 0.89                              & 4.56             & 0.77           & \textbf{3.95}      & \textbf{0.07}     \\ \hline
Boston                            & 4.39                & 0.71               & 4.57                   & 0.64                  & 5.03                               & 0.59                              & 5.11             & 0.54           & \textbf{2.75}      & \textbf{0.06}     \\ \hline
Auto MPG                          & 4.24                & 1.01               & 4.35                   & 0.29                  & 5.11                               & 0.31                              & 6.27             & 0.36           & \textbf{2.81}      & \textbf{0.10}     \\ \hline
Energy Appliance                  & 87.27               & 0.21               & 86.87                  & 0.33                  & 88.37                              & 1.03                              & 86.92            & 0.91           & \textbf{86.17}     & \textbf{0.13}     \\ \hline
Superconductivity                 & 10.97               & 0.49               & 11.12                  & 0.57                  & 12.33                              & 0.61                              & 10.82            & 2.31           & \textbf{10.79}     & \textbf{0.11}     \\ \hline
\end{tabular}
\label{table:perf}
\end{table*}

\begin{figure*}[t]
\centering
\subfigure[Parkinsons]{\includegraphics[width=0.24\linewidth]{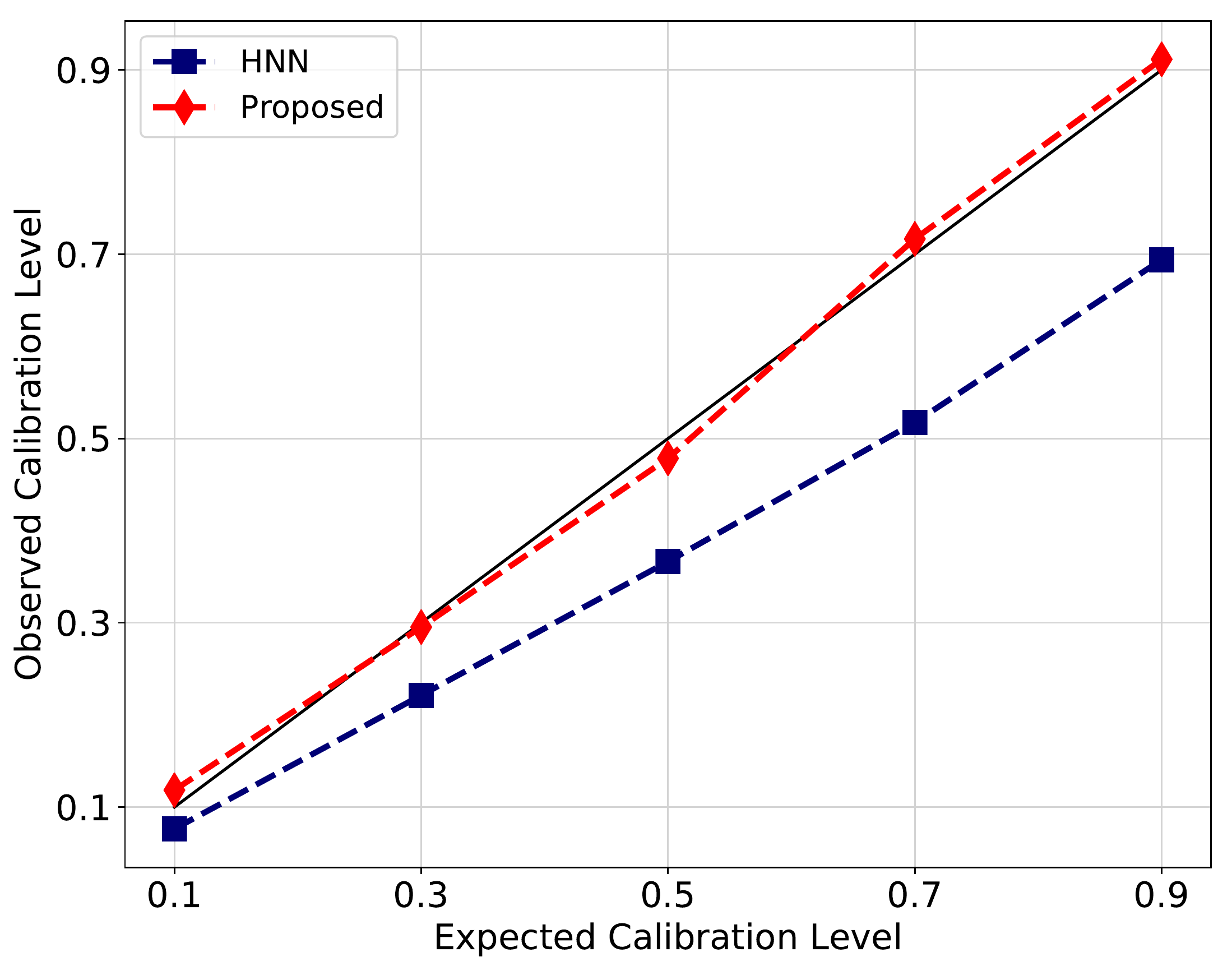}}
\subfigure[Red Wine]{\includegraphics[width=0.24\linewidth]{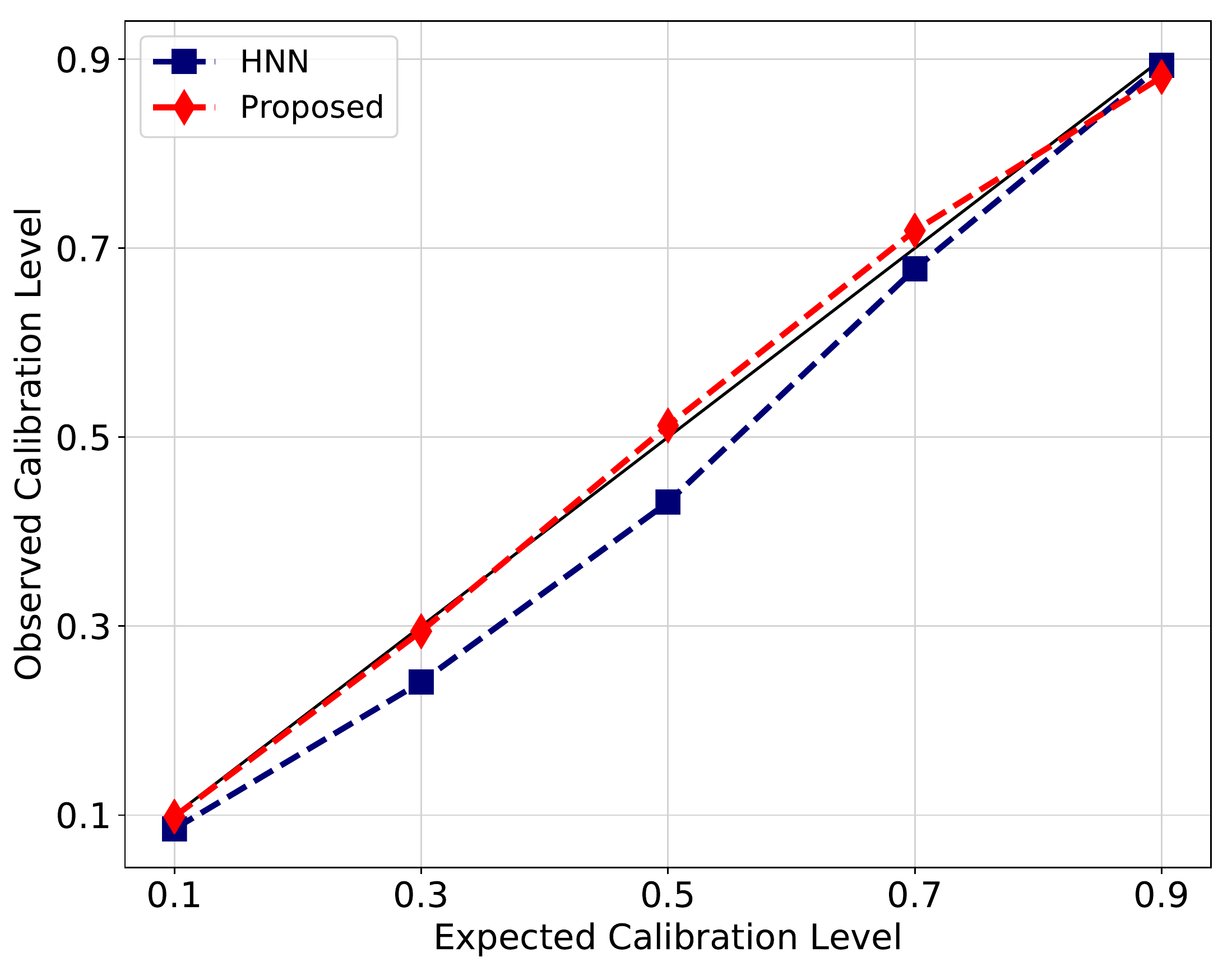}}
\subfigure[Energy Appliance]{\includegraphics[width=0.24\linewidth]{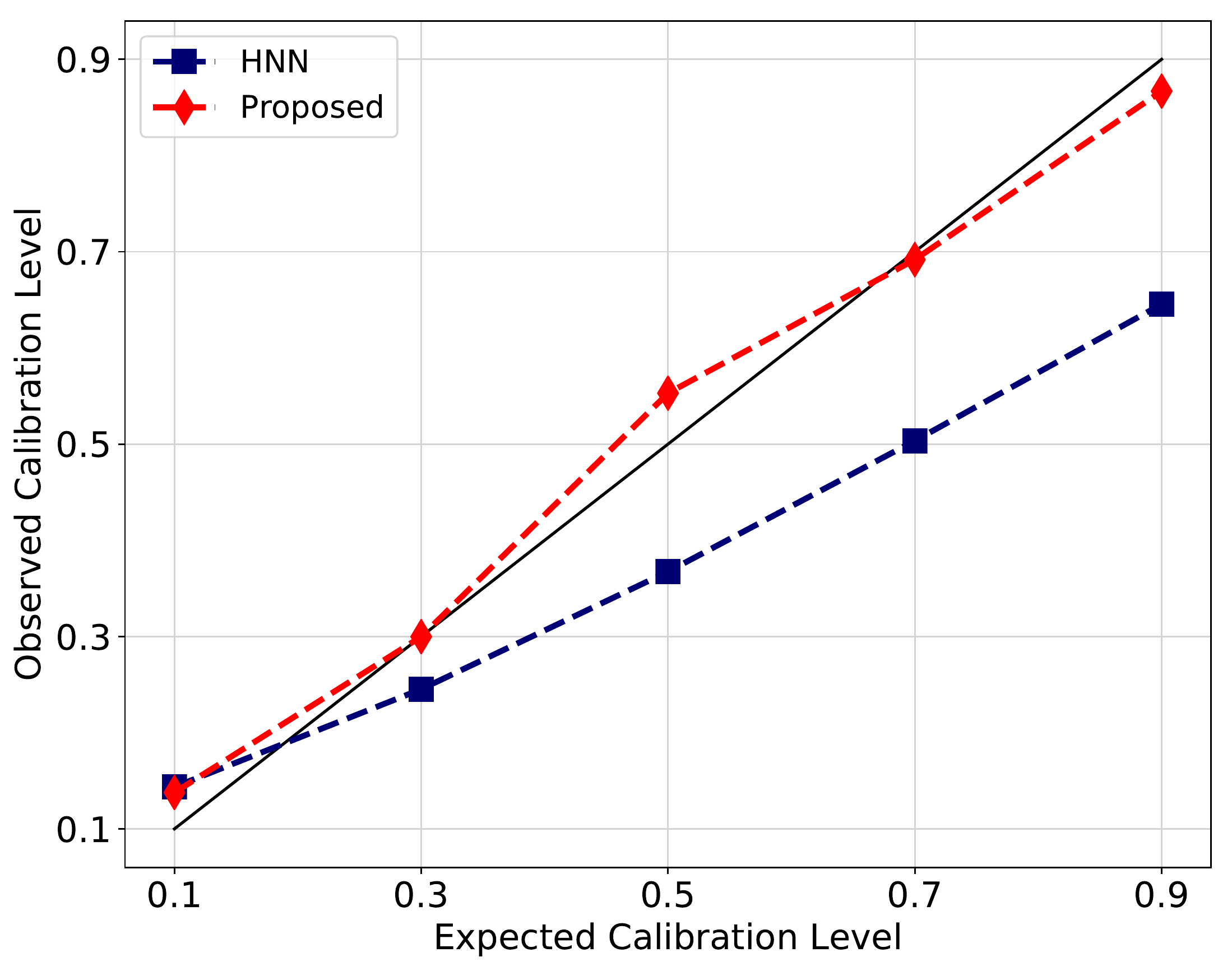}}
\subfigure[Auto MPG]{\includegraphics[width=0.24\linewidth]{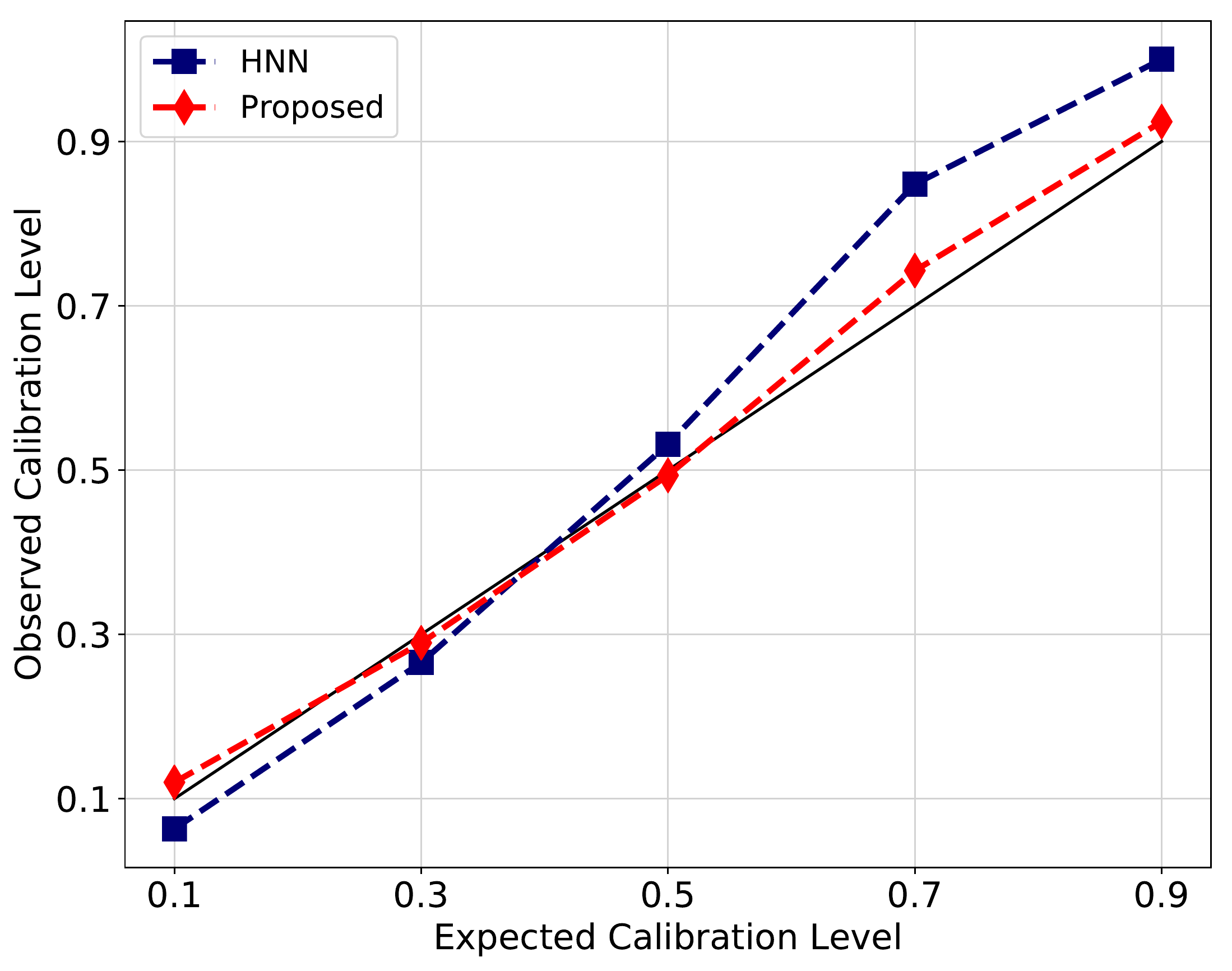}}
\caption{Calibration plots for the proposed approach on different benchamark datasets. For comparison, we show the results from another black-box estimator, heteroscedastic neural networks.}
\label{fig:calib}
\end{figure*}

Our algorithm employs an alternating optimization strategy that updates $\Theta$ and $\Phi$ with the goal of improving calibration for the chosen set of  $\alpha$'s. In particular, we use an empirical calibration error metric for refining the model $\mathcal{G}$:
\begin{align}
\label{eqn:ece}
&\Phi^* = \arg \min_{\Phi} \mathrm{L}_{\mathcal{G}} \\
\nonumber&= \arg \min_{\Phi} \sum_{\alpha \in A} \bigg(\left|\alpha - \frac{1}{N}{ \sum_{i=1}^{N}\mathbb{I}\left[\hat{\mathrm{y}}_i - \mathrm{\delta}_i^{\alpha} \leq \mathrm{y}_i \leq \hat{\mathrm{y}}_i + \mathrm{\delta}_i^{\alpha}\right]} \right| \\
\nonumber & \quad \quad \quad \quad \quad \quad + \lambda_1 \left|(\hat{\mathrm{y}}_i + \mathrm{\delta}_i^{\alpha}) - \mathrm{y}_i\right| + \lambda_2 | \mathrm{y}_i - (\hat{\mathrm{y}}_i - \mathrm{\delta}_i^{\alpha})|\bigg).
\end{align}Here $A$ indicates the set of calibration levels that we want to simultaneously achieve using the corresponding width estimates $\mathrm{\delta}^{\alpha}$ and $N$ denotes the total number of samples. This metric measures the discrepancy between the true  empirical probability and the likelihood of the true target falling in the interval. Note that the mean estimates $\hat{\mathrm{y}}_i = \mathcal{F}(\mathrm{x}_i; \Theta)$ are obtained using the current state of the parameters $\Theta$. The last two terms are used as regularizers to penalize larger widths so that trivial solutions are avoided. The hyperparameters $\lambda_1$ and $\lambda_2$ are set to $0.1$ in all our experiments. In practice, we find that such a simultaneous optimization is challenging and the loss function is biased towards larger values of $\alpha$. Hence, in our algorithm, we randomly choose a level $\alpha$ from the set $A$ in each iteration. 

The key idea of the proposed approach is to leverage the estimated intervals to drive the update of the mean estimator. To this end, we propose to employ a hinge loss objective that attempts to adjust the mean estimate such that the observed likelihood of the target contained in the interval increases:
\begin{align}
\label{eqn:hinge}
&\Theta^* = \arg \min_{\Theta} \mathrm{L}_{\mathcal{F}} \\
\nonumber&= \arg \min_{\Theta} \sum_{i=1}^N w_i \bigg[\max(0, (\hat{\mathrm{y}}_i - \mathrm{\delta}_i^{\alpha}) - \mathrm{y}_i +  \tau)\\ 
\nonumber & \quad \quad \quad \quad \quad \quad + \max(0, \mathrm{y}_i  - (\hat{\mathrm{y}}_i + \mathrm{\delta}_i^{\alpha}) + \tau)\bigg].
\end{align}Here, $\mathrm{\delta}_i^{\alpha}$ is obtained using the current state of the estimator $\mathcal{G}$ for the $\alpha$ chosen during that iteration. The optional threshold $\tau$ is set to $0.05$ in all our experiments. The weights $w_i = \mathrm{\delta}_i^{\alpha}/\sum_j \mathrm{\delta}_j^{\alpha}$ penalize samples with larger widths (less confident) while updating $\Theta$. Intuitively, for a fixed interval, the improved mean estimate can potentially increase the calibration error by achieving a higher likelihood even for smaller $\alpha$ levels. However, in the subsequent step of updating $\Phi$, we expect the widths to become sharper in order to reduce the calibration error. As a result, this collaborative optimization process leads to superior quality mean estimates and highly calibrated intervals. Though a rigorous analysis of the intervals remains to be done, the value of using calibration as a training objective is clearly evident in our experiments.

\section{Performance Evaluation}
\label{sec:results}
\begin{figure*}[t]
\centering
\subfigure[]{\includegraphics[width=0.24\linewidth]{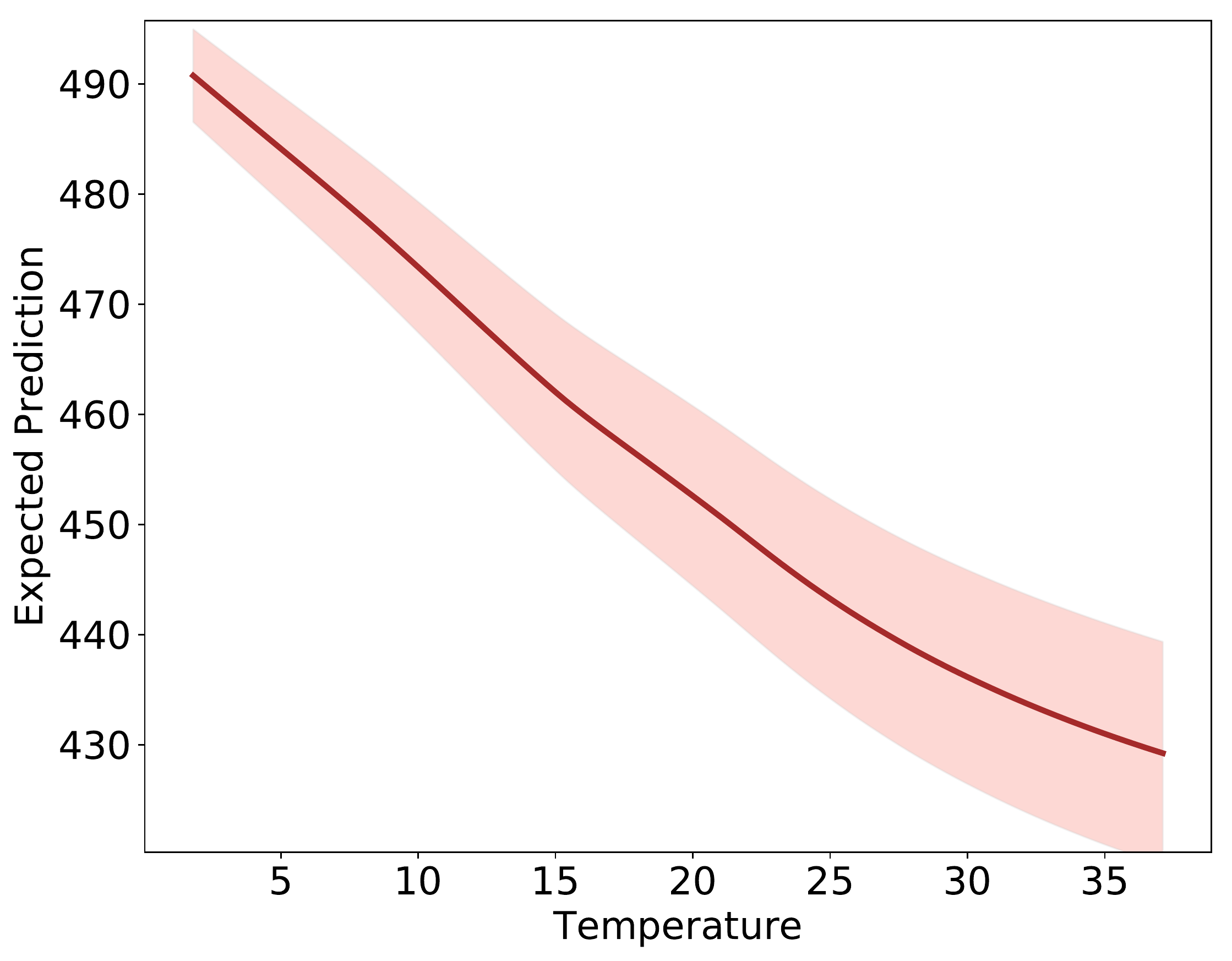}}
\subfigure[]{\includegraphics[width=0.24\linewidth]{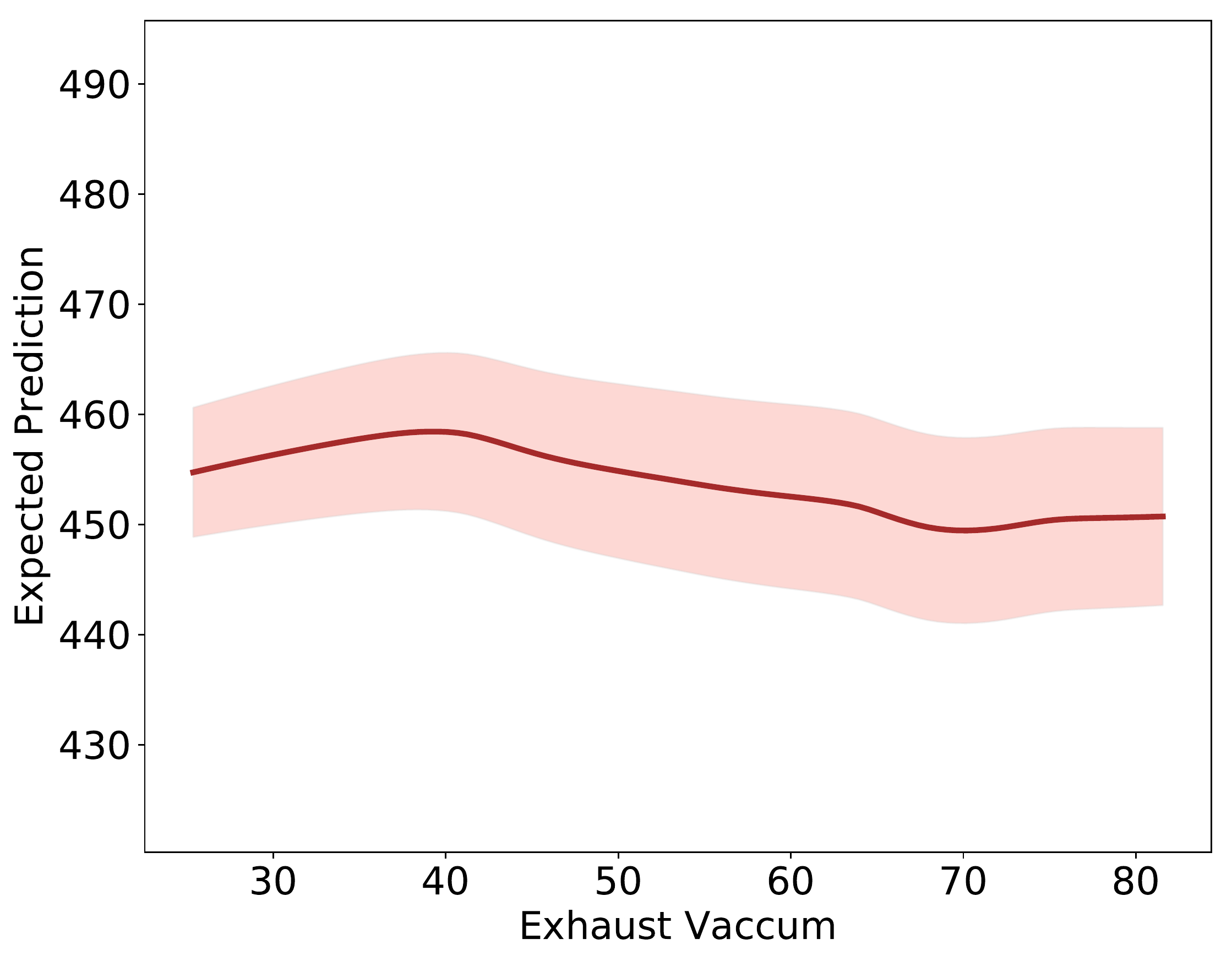}}
\subfigure[]{\includegraphics[width=0.24\linewidth]{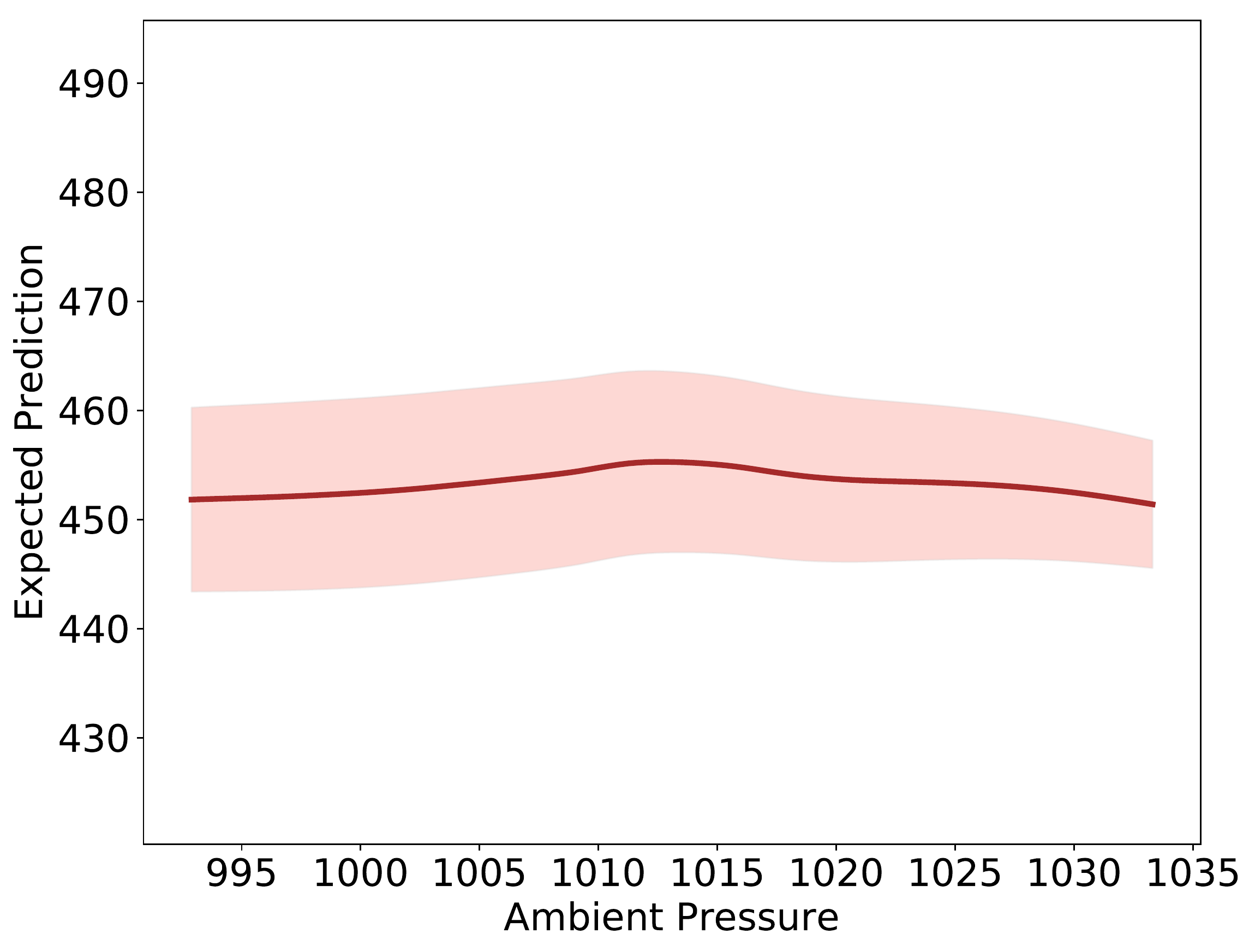}}
\subfigure[]{\includegraphics[width=0.24\linewidth]{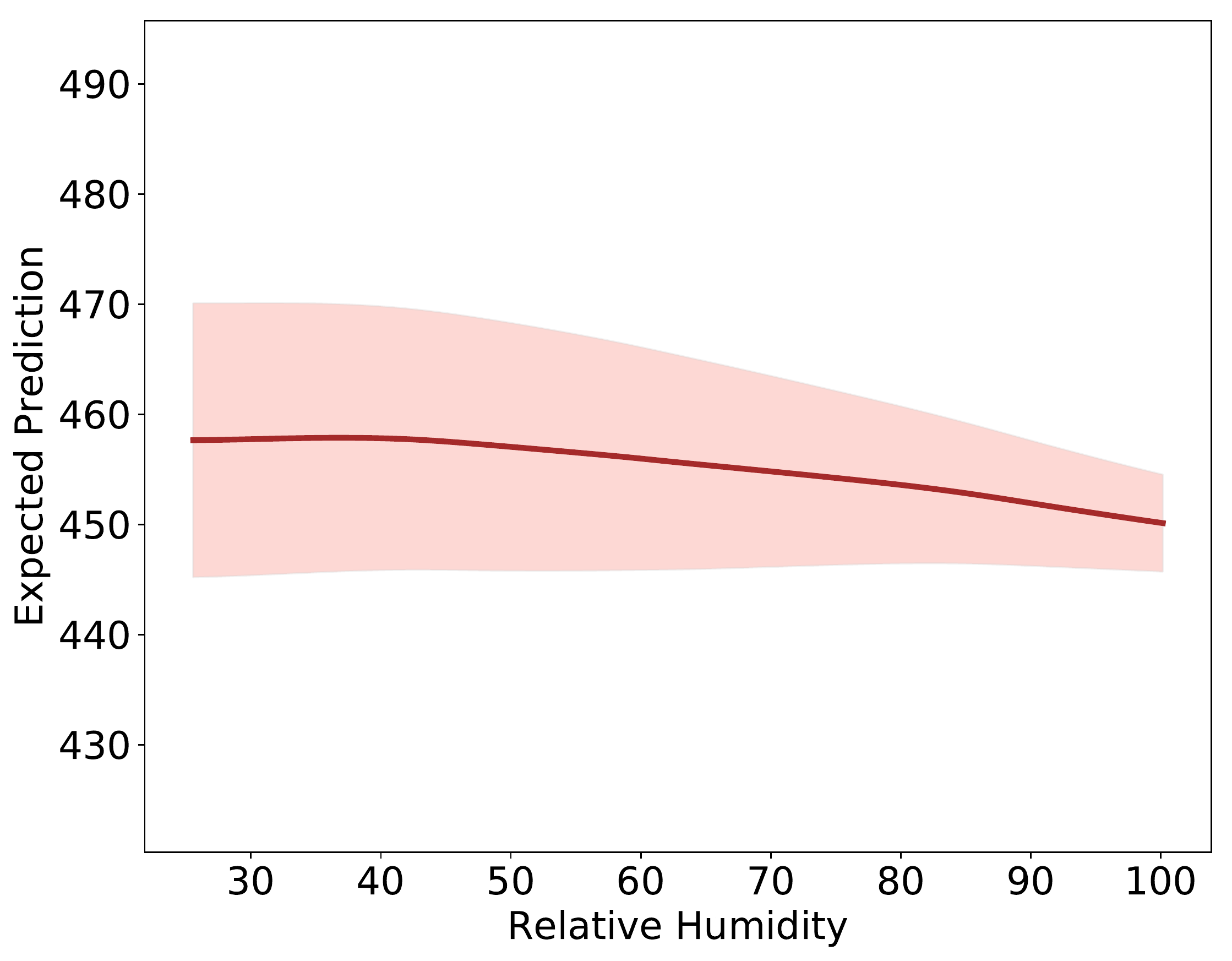}}
\caption{\textit{Power plant dataset} - Partial dependence analysis of the $4$ input variables on the net energy output.}
\label{fig:cccp}
\end{figure*}
We evaluate the proposed approach using benchmark regression tasks and compare its performance against existing baseline methods for deep predictive modeling. We consider $8$ different datasets from the UCI repository~\cite{Dua:2019} and use random 80-20 splits for training and testing respectively. We performed $5-$fold cross validation and measured the average performance in all cases. All experiments were carried out using a neural network with $5$ fully connected layers with ReLU non-linearity, and a final regression layer. Following standard practice we report the root mean squared error (RMSE) and the empirical calibration error (ECE) metrics for evaluation. For the ECE metric, we used the set $A = [0.1, 0.3, 0.5, 0.7, 0.9]$. We used the following hyper-parameters in our experiments: $\lambda_1 = \lambda_2 = 0.1$, $T = 1000$, learning rates of $5e-5$ and $1e-4$ for updating the parameters $\Theta$ and $\Phi$ respectively. For comparison, we considered the following baseline methods: (i) MC dropout~\cite{gal2016dropout}, (ii) concrete dropout~\cite{gal2017concrete}, (iii) Bayesian neural networks (BNN)~\cite{ghahramani2015probabilistic} and (iv) heteroscedastic neural networks~\cite{gal2016uncertainty}. Note that, these approaches include the uncertainty estimation step as part of the training process and have varying degrees of impact on the behavior of the resulting mean and interval estimators.

\begin{figure}[t]
\centering
\subfigure[Age]{\includegraphics[width=0.48\linewidth]{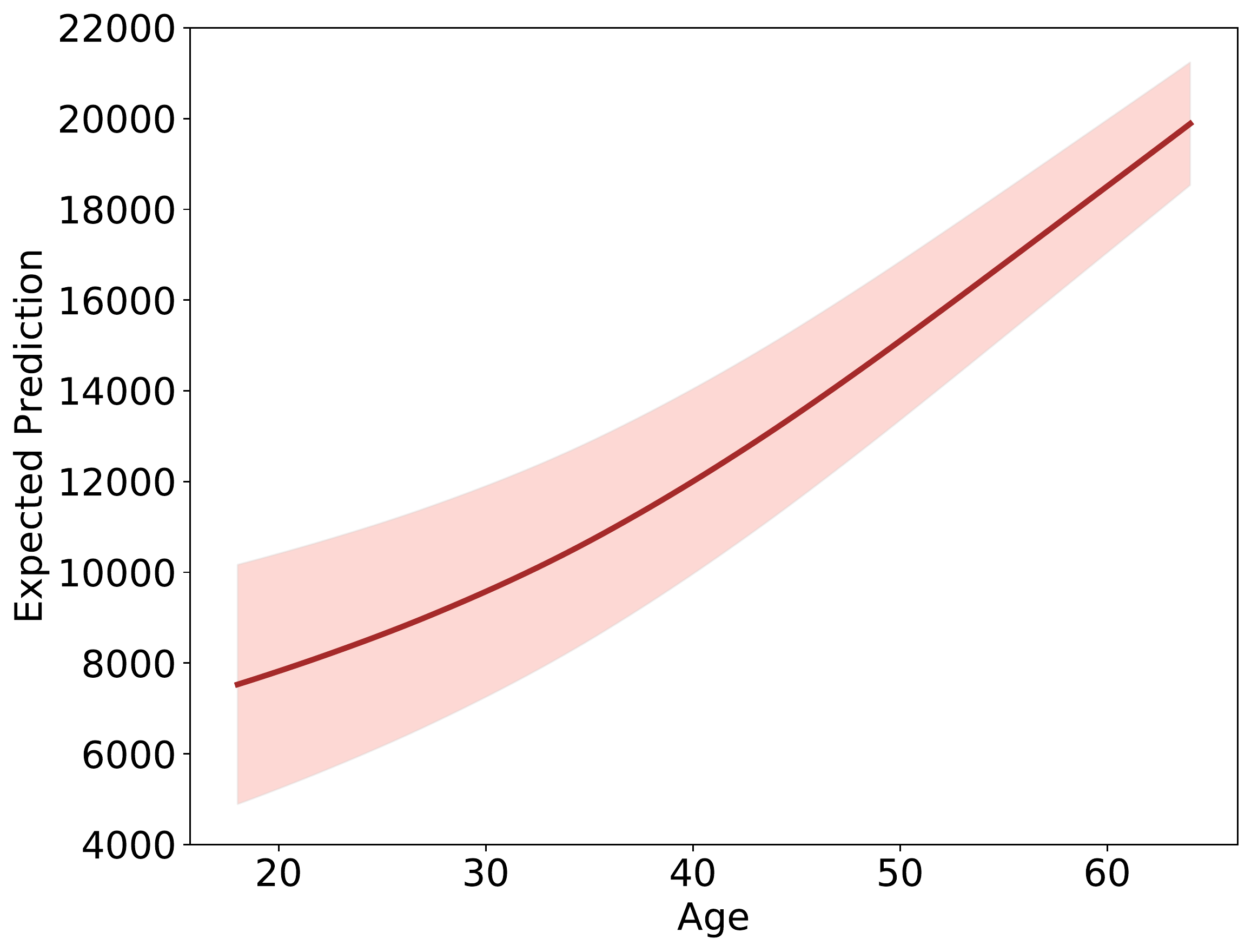}}
\subfigure[BMI]{\includegraphics[width=0.48\linewidth]{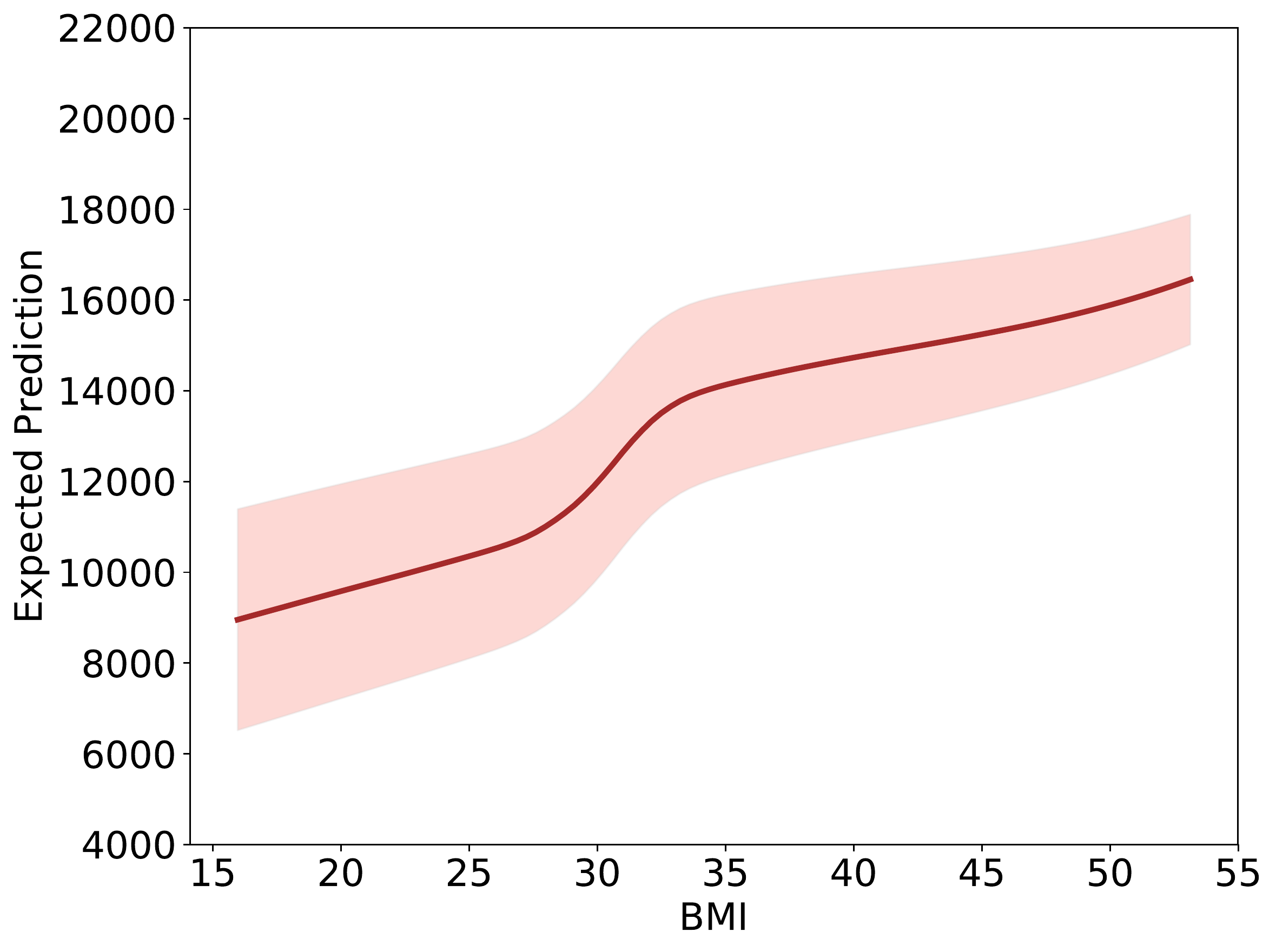}}
\caption{\textit{Insurance cost prediction} - Partial dependence analysis of \textit{Age} and \textit{BMI} variables. Here, we augment the expected predictions in PD plots with interval estimates.}
\label{fig:insurance}
\end{figure}

As seen in Table~\ref{table:perf}, the conventional dropout regularization produces high-fidelity predictors in terms of test RMSE, but produces unsatisfactory calibration error in many cases. In contrast, the Bayesian neural network method, which utilizes a variational inferencing approach for incorporating model uncertainties into the learning, produces consistently better calibration, while achieving larger RMSE. In comparison, concrete dropout and HNN demonstrate a better trade-off between the two metrics. The proposed approach improves significantly over the baselines, in terms of both the metrics, and clearly evidences the power of calibration as a learning objective. Though being a black-box interval estimation method (without any explicit uncertainty estimation step), similar to HNN, the proposed approach matches the empirical probabilities at different confidence levels (see calibration plots in Figure~\ref{fig:calib}).

\section{Analysis}
\label{sec:pda}
Gaining insights into a model's behavior is critical to its deployment and usage in the real-world. Consequently, it is common to utilize model-agnostic interpretation tools to explore the characteristics of a learned model. In regression models, the partial dependence plot (PDP)~\cite{friedman2001greedy} is a widely adopted tool for studying the marginal effect of each (or two) feature on the predicted outcome of a model. PDP reveals the global relationship between the target and a feature, for example if it is linear or monotonic. Assuming $\mathrm{x}^s$ to denote the feature for which the PDP is plotted, and $\mathrm{x}^c$ to denote the other $d-1$ features, such that $\mathrm{x} = [\mathrm{x}^s, \mathrm{x}^c]$ the partial dependence function can be evaluated on the training data as:

\begin{equation}
    \mathrm{P}(\mathrm{x}^s) = \frac{1}{N_t} \sum_{i = 1}^{N_t} \mathcal{F}(\mathrm{x}^s, \mathrm{x}_i^c).
\end{equation}Here $N_t$ denotes the total number of training samples. In this formula, $\mathrm{x}_i^c$ corresponds to actual features from observed data, and the partial function $\mathrm{P}$ is evaluated for each value of $\mathrm{x}^s$. We propose to augment PD plots with expected intervals, in order to obtain a better understanding of the dependencies. For example in Figure~\ref{fig:insurance}, we show the PDP for the \textit{Age} and \textit{BMI variables} on the insurance cost, wherein the dataset contains additional variables including age, gender and medical history. In the augmented PD plot for \textit{Age}, we can observe from the mean prediction that the cost monotonically increases. Further, from the expected interval estimates, we see that, at lower \textit{Age} values, say $20$, the intervals are large enough to include the expected costs even at age $35$. However, as the \textit{Age} variable grows, the intervals are sharp indicating that predictions are highly sensitive in that regime. On the other hand, the \textit{BMI} variable shows a clear split around the value $35$, with minimal variations (less sensitive) within the two regions. Similarly, the PD plots for the $4$ variables in the UCI power plant dataset~\cite{Dua:2019} reveal that the \textit{Temperature} parameter has a strong dependency (inverse) with the energy output, and is significantly more sensitive at lower values. On the other hand, though the other $3$ parameters show no apparent relationship, the intervals for the PDP of \textit{Relative Humidity} reveals a more complex relationship. In summary, we find calibration to be effective for building predictive models, and the resulting intervals can be useful in practice, even though they cannot be directly associated to specific uncertainties.

\bibliographystyle{IEEEbib}
\bibliography{ref}

\end{document}